%% file: starSEM.tex
\documentclass[11pt,a4paper,spanish]{article}
\usepackage[hyperref]{naaclhlt2019}
\usepackage{times}
\usepackage{latexsym} 
\usepackage{algorithm}
\usepackage[noend]{algpseudocode}
\usepackage{mathtools}
\usepackage{acronym}
\usepackage{graphicx}
\usepackage{selinput}
\SelectInputMappings{%
  aacute={á},
  ntilde={ñ},
  Euro={€}
}
\usepackage{babel}

\usepackage{comment}
\usepackage{enumitem}
\usepackage{mathrsfs,amsmath}
\usepackage[normalem]{ulem}
\usepackage{amsmath}

\usepackage{url}
\usepackage{array}

\aclfinalcopy 

\makeatletter
\newcommand*{\rom}[1]{\expandafter\@slowromancap\romannumeral #1@}
\makeatother
\DeclareMathOperator*{\argmin}{arg\,min}
\DeclarePairedDelimiter{\abs}{\lvert}{\rvert}

\title{Generating Animations from Screenplays}

\author{ Yeyao Zhang$^{1,3}$, 
  Eleftheria Tsipidi$^{1}$ , \\ 
  \textbf{Sasha Schriber$^{1}$,} 
  \textbf{Mubbasir Kapadia$^{1,2}$,} 
  \textbf{Markus Gross$^{1,3}$},
  \textbf{Ashutosh Modi$^{1}$} \\ 
  {$^1$Disney Research, } 
  {$^{2}$Rutgers University, }
  {$^{3}$ETH Zurich}\\
  {\tt yezhang@inf.ethz.ch}  \\
  {\tt \{etsipidi,sasha.schriber\}@disneyresearch.com}\\
  {\tt mubbasir.kapadia@rutgers.edu} \\
  {\tt \{gross,ashutosh.modi\}@disneyresearch.com}
\\}

\date{}

\begin{document}
\maketitle
\begin{abstract}

Automatically generating animation from natural language text finds application in a number of areas e.g. movie script writing, instructional videos, and public safety. However, translating natural language text into animation is a challenging task. Existing text-to-animation systems can handle only very simple sentences, which limits their applications. 
In this paper, we develop a text-to-animation system which is capable of handling complex sentences. We achieve this by introducing a text simplification step into the process. Building on an existing animation generation system for screenwriting, we create a robust NLP pipeline to extract information from screenplays and map them to the system's knowledge base. We develop a set of linguistic transformation rules that simplify complex sentences. Information extracted from the simplified sentences is used to generate a rough storyboard and video depicting the text. Our sentence simplification module outperforms existing systems in terms of BLEU and SARI metrics.We further evaluated our system via a user study: 68\% participants believe that our system generates reasonable animation from input screenplays.

\end{abstract}

\input{introduction}
\input{relatedWork}
\input{systemArchitecture}
\input{data}
\input{evaluation}
\input{conclusion}

\bibliographystyle{acl_natbib}
\bibliography{starSEM}

\appendix
\input{Appendix/appendixa.tex}{}
\clearpage
\appendix
\input{Appendix/appendixb.tex}{}
\clearpage
\appendix
\input{Appendix/appendixc.tex}{}

\end{document}

%% file: introduction.tex
\section{Introduction}

Generating animation from texts can be useful in many contexts e.g. movie script writing \cite{Ma:2006, Liu2006, Hanser2009}, instructional videos \cite{lu2002automatic}, and public safety \cite{johansson2004carsim}. Text-to-animation systems can be particularly valuable for screenwriting by enabling faster iteration, prototyping and proof of concept for content creators. 

In this paper, we propose a text-to-animation generation system. Given an input text describing a certain activity, the system generates a rough animation of the text. We are addressing a practical setting, where we do not have any annotated data for training a supervised end-to-end system. The aim  is not to generate a polished, final animation, but a \textit{pre-visualization} of the input text. The purpose of the system is not to replace writers and artists, but to make their work more efficient and less tedious. We are aiming for a system which is robust and could be deployed in a production environment.  

Existing text-to-animation systems for screenwriting (\textsection \ref{sec:Related_Work}) visualize stories by using a pipeline of Natural Language Processing (NLP) techniques for extracting information from texts and mapping them to appropriate action units in the animation engine. The NLP modules in these systems translate the input text into predefined intermediate action representations and the animation generation engine produces simple animation from these representations. 

Although these systems can generate animation from carefully handcrafted simple sentences, translating real screenplays into coherent animation still remains a challenge. This can be attributed to the limitations of the NLP modules used with regard to handling complex sentences. In this paper, we try to address the limitations of the current text-to-animation systems. Main contributions of this paper are:
\begin{itemize}[noitemsep,topsep=0pt]
    \item We propose a screenplay parsing architecture which generalizes well on different screenplay formats (\textsection \ref{sec:parsing}).
    \item We develop a rich set of linguistic rules to reduce complex sentences into simpler ones to facilitate information extraction (\textsection \ref{sec:simplify}).
    \item We develop a new NLP pipeline to generate animation from actual screenplays (\textsection \ref{sec:Sys_arch}). 
\end{itemize}

\begin{figure*}[t]
\centering
  \includegraphics[width=\textwidth]{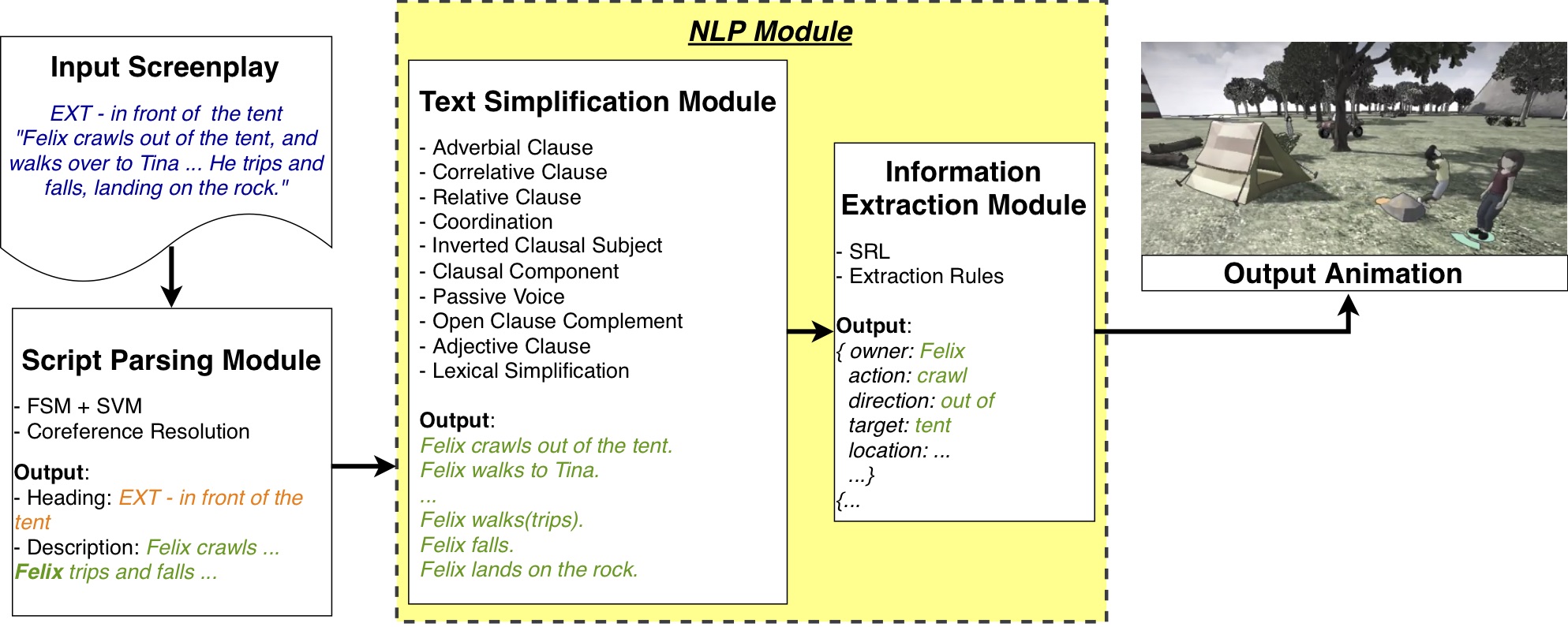}\hfill
  \caption{ System Architecture: Screenplays are first segmented into different functional blocks. Then, the descriptive action sentences are simplified. Simplified sentences are used to generate animation.} 
  \label{fig:overview}
\end{figure*}


The potential applications of our contributions are not restricted to just animating screenplays. The techniques we develop are fairly general and can be used in other applications as well e.g. information extraction tasks. 

%% file: relatedWork.tex
\section{Related Work}\label{sec:Related_Work}

Translating texts into animation is not a trivial task, given that neither the input sentences nor the output animations have a fixed structure. Prior work addresses this problem from different perspectives \cite{Hassani2016}. 

CONFUCIUS \cite{Ma:2006} is a system that converts natural language to animation using the FDG parser \cite{fdg} and WordNet \cite{wordnet}. ScriptViz \cite{Liu2006} is another similar system, created for screenwriting. It uses the Apple Pie parser \cite{Sekine} to parse input text and then recognizes objects via an object-specific reasoner. It is limited to sentences having conjunction between two verbs. SceneMaker \cite {Hanser2009} adopts the same NLP techniques as proposed in CONFUCIUS \cite{Ma:2006} followed by a context reasoning module. 
Similar to previously proposed systems, we also use dependency parsing followed by linguistic reduction (\textsection \ref{sec:simplify}). 

Recent advances in deep learning have pushed the state of the art results on  different NLP tasks \cite{spacy-parser,neuralcoref,he2017deep}. 
We use pre-trained models for dependency parsing, coreference resolution and SRL to build a complete NLP pipeline to create intermediate action representations. For the action representation (\textsection \ref{sec:srl}), we use a key-value pair structure inspired by the PAR architecture \cite{PAR}, which is a knowledge base of representations for actions performed by virtual agents.

Our work comes close to the work done in the area of Open Information Extraction (IE) \cite{niklaus2018survey}. In particular, to extract information, Clause-Based Open IE systems \cite{del2013clausie, angeli2015leveraging, schmidek2014improving} reduce a complex sentence into simpler sentences using linguistic patterns. However, 
the techniques developed for these systems do not generalize well to screenplay texts, as these systems have been developed using well-formed and factual texts like Wikipedia, Penn TreeBank, etc. An initial investigation with the popular Open IE system OLLIE (Open Language Learning for Information Extraction) \cite{mausam2012open} did not yield good results on our corpus.

Previous work related to information extraction for narrative technologies includes the CARDINAL system \cite{marti2018CARDINAL,sanghrajka2018computer}, as well as the conversational agent PICA \cite{falk2018pica}. They focus on querying knowledge from stories. The CARDINAL system also generates animations from input texts. However, neither of the tools can handle complex sentences. We build on the CARDINAL system. We develop a new NLP module to support complex sentences and leverage the animation engine of CARDINAL.

Recently, a number of end-to-end image generation systems have been proposed \cite{mansimov2015generating,reed2016generative}. But these systems do not synthesize satisfactory images yet, and are not suitable for our application.  It is hoped that the techniques proposed in this paper could be used for automatically generating labelled data (e.g. (text,video) pairs) for training end-to-end text-to-animation systems.

%% file: systemArchitecture.tex
\section{Text-to-Animation System} \label{sec:Sys_arch}

We adopt a modular approach for generating animations from screenplays. The general overview of our approach is presented in Figure~\ref{fig:overview}. The system is divided into three modules:
\begin{itemize}[noitemsep,topsep=0pt]
\item \textbf{Script Parsing Module:} Given an input screenplay text, this module  automatically extracts the relevant text for generating the animation (\textsection\ref{sec:parsing}). 
\item \textbf{NLP Module:} It processes the extracted text to get relevant information. This has two sub-modules: 

\begin{itemize}[noitemsep,topsep=0pt]
    \item \textbf{Text Simplification Module:} It simplifies complex sentences using a set of linguistic rules (\textsection\ref{sec:simplify}). 
    \item \textbf{Information Extraction Module:} It extracts information from the simplified sentences into pre-defined action representations (\textsection\ref{sec:srl}).
\end{itemize}
\item\textbf{Animation Generation Module:} It generates animation based on action representations  (\textsection\ref{sec:animation}).
\end{itemize}

\subsection{Script Parsing Module}
\label{sec:parsing}
Typically, \textit{screenplays} or \textit{movie scripts} or \textit{scripts} (we use the terms interchangeably), are made of several scenes, each of which corresponds to a series of consecutive motion shots. Each scene contains several functional components\footnote{\url{https://www.storysense.com/format.htm}}: \textit{Headings} (time and location), \textit{Descriptions} (scene description, character actions), \textit{Character Cues} (character name before dialog), \textit{Dialogs} (conversation content), \textit{Slug Lines} (actions inserted into continuous dialog) and \textit{Transitions} (camera movement). In many scripts, these components are easily identifiable by indentation, capitalization and keywords. We call these scripts \textit{well-formatted}, and the remaining ones \textit{ill-formatted}. We want to segment the screenplays into components and are mainly interested in the Descriptions component for animation generation.

\begin{algorithm}[t]
    \scriptsize	\caption{Syntactic Simplification Procedure}\label{alg:syntac-sys}
	\begin{algorithmic}[1]
		\Procedure{Syntactic Simplification}{$sent, temp$}
		\State Q $\gets$ empty queue
		\State HS $\gets$ empty integer hash set
		\State RES $\gets$ empty list
		\State Q.push(sent) \Comment{push input sentence to queue}
		\While{Q $\ne$ Empty}
		\State str $\gets$ Q.pop()
		\If{hash(str) $\in$ HS}
		\State RES.append(str)\Comment{have seen this sentence}
		\State \textbf{continue}
		\EndIf
		\State HS.add(str) \Comment{mark current sentence as \textit{already seen}}
		\State transform $\gets$ False
		\For{a in analyzers}
		\If{!transform $\&$ a\textit{.identify}(str)}
		\State transform $\gets$ True
		\State \emph{simplified} = a\textit{.transform}(str)
		\State correct\_verb\_tense(\emph{simplified})
		\State Q.push(\emph{simplified})
		\EndIf
        
		\EndFor
        
		\If{transform $\ne$ True}
		\State RES.append(str)
		\EndIf
		\EndWhile
        
		\State \textbf{return} RES
		\EndProcedure
	\end{algorithmic}

\end{algorithm}

\noindent\textbf{Well-formatted Scripts:} We initially tried  ScreenPy~\cite{Screenpy} to annotate the well-formatted scripts with the component labels. However, ScreenPy did not perform well on our data. We developed our own model, based on Finite State Machines (FSM), for parsing scripts (for details refer to Appendix A). Due to space limitations, we do not describe the FSM model; the key idea is that the model uses hand-crafted transition rules to segment the input screenplay and generates (paragraph, component name) pairs.


\noindent\textbf{Ill-formatted Scripts:}
Taking all the (paragraph, component name) pairs generated by the FSM as ground truth, an SVM model is trained to segment ill-formatted screenplays with inconsistent indentations. For extracting features, each paragraph is encoded into a fix-sized vector using a pre-trained Universal Sentence Encoder. 
The SVM is trained and tested on a 9:1 train-test data split. The result shows an accuracy of 92.72\% on the test set, which is good for our purposes, as we are interested mainly in the Descriptions component.

\begin{table*}[t]
	\centering \scriptsize
	\begin{tabular}{|p{0.20\linewidth}|p{0.30\linewidth}|p{0.40\linewidth}|}
	\hline
	Syntactic Structure&\emph{Identify} procedure&\emph{Transform} procedure\\
	\hline
	Coordination&
	search if \emph{cc} and \emph{conj} in dependency tree&
	cut \emph{cc} and \emph{conj} link. If \emph{conj} is verb, mark it as new root; else replace it with its sibling node.\\
	\hline
	Pre-Correlative Conjugation&
    locates position of keywords: ``either'', ``both'',``neither''&
	removed the located word from dependency tree\\
	\hline
	Appositive Clause&
	find \emph{appos} token and its head (none)&
	glue appositive noun phrase with ``to be''\\
	\hline
	Relative Clause&
	find \emph{relcl} token and its head&
	cut \emph{appos} link, then traverse from root. Then, if no ``wh'' word present, put head part after anchor part; else, we divide them into 5 subcases (Table \ref{tab:simp-res})\\
	\hline
	Adverbial Clause Modifier&
	find \emph{advcl} token and its head. Also conjuncts of head token&
	cut \emph{advcl} edge. If \emph{advcl} token does not have subject, add subject of root as \emph{advcl}'s most-left child and remove \emph{prep} and \emph{mark} token. Then traverse from both root and \emph{advcl} token\\
	\hline
	Inverted Clausal Subject&
	\emph{attr} token has to be the child of head of \emph{csubj} token&
	change position of actual verb and subject\\
	\hline
	Clausal Complement&
	find \emph{ccomp} token in dependency tree&
	cut \emph{ccomp} link, add subject to subordinate clause if necessary\\
	\hline
	Passive Voice&
	check presence of \emph{nsubjpass} or \emph{csubjpass} optionally for \emph{auxpass} and \emph{agent}&
	cut \emph{auxpass} link if any. Cut \emph{nsubjpass} or \emph{csubjpass} link. Prepend subject token to verb token's right children. Finally append suitable subject.\\
	\hline
	Open Clause Complement&
	find \emph{xcomp} verb token adn actual verb token&
	if \emph{aux} token presents, cut \emph{aux} link, then replace xcomp-verb in subject's children with actual-verb, traverse from actual-verb; else, cut \emph{xcomp} link, traverse from xcomp-verb\\
	\hline
	Adjective Clause&
	find \emph{acl} verb token and its head&
	cut \emph{acl} link. Link subject node with it. Traverser from \emph{acl} node\\
	\hline
	\end{tabular}
    \caption[]{Linguistic rules for text simplification module \label{tab:simp-rules}}
\end{table*}

\noindent\textbf{Coreference Resolution:}
Screenplays contain a number of ambiguous entity mentions (e.g. pronouns). In order to link mentions of an entity, an accurate coreference resolution system is required. The extracted Descriptions components are processed with the NeuralCoref\footnote{\url{github.com/huggingface/neuralcoref}} system. Given the text, it resolves mentions (typically pronouns) to the entity they refer to in the text.  To facilitate entity resolution, we prepend each Description component with the Character Cues component which appears before it in the screenplay (e.g. [character]MARTHA: [dialog]``I knew it!'' [description]She then jumps triumphantly $\rightarrow{}$ MARTHA. She then jumps triumphantly).



\subsection{Text Simplification Module}
\label{sec:simplify}
In a typical text-to-animation system, one of the main tasks is to process the input text to extract the relevant information about actions (typically verbs) and participants (typically subject/object of the verb), which is subsequently used for generating animation. This works well for simple sentences having a single verb with one subject and one (optional) object. However, the sentences in a screenplay are complicated and sometimes informal. In this work, a sentence is said to be complicated if it deviates from easily extractable and simple subject-verb-object (and its permutations) syntactic structures 
and possibly has multiple actions mentioned within the same sentence with syntactic interactions between them. By syntactic structure we refer to the dependency graph of the sentence. 

In the case of screenplays, the challenge is to process such complicated texts. We take the text simplification approach, i.e. the system first simplifies a complicated sentence and then extracts the relevant information. Simplification reduces a complicated sentence into multiple simpler sentences, each having a single action along with its participants, making it straightforward to extract necessary information. 

Recently, end-to-end Neural Text Simplification (NTS) systems ~\cite{nisioi2017exploring,saggion2017automatic} have shown reasonable accuracy. However, these systems have been trained on factual data such as Wikipedia and do not generalize well to screenplay texts. Our experiments with such a pre-trained neural text simplification system did not yield good results (\textsection \ref{sec:intrinsic}). Moreover, in the context of text-to-animation, there is no standard labeled corpus to train an end-to-end system. 

\begin{table*}[t]
	\centering \scriptsize
	\begin{tabular}{|p{0.13\linewidth}|p{0.30\linewidth}|p{0.23\linewidth}|p{0.23\linewidth}|}
	\hline
		Type&Example Input Sentence&System Output Sentence 1&System Output Sentence 2\\
		\hline
		Coordination&She LAUGHS, and[\textbf{cc}] gives[\textbf{conj}] Kevin a kiss.&She laughs.&She gives Kevin a kiss.\\
		\hline
		Pre-Correlative& It's followed by another squad car, both[\textbf{preconj}] with sirens blaring. & It's followed by another squad car, with sirens blaring.&--\\
		\hline
		Appositive&Kevin is reading a book the Bible[\textbf{appos}]&Kevin reads a book&The book is the Bible.\\
		\hline
		Relative-dobj&She pulls out a letter \emph{which}[\textbf{dobj}] she hands[\textbf{relcl}] to Keven&Shee pulls out a letter&She hands a lettre to Kevin.\\
		\hline
		Relative-pobj&A reef encloses the cove \emph{where}[\textbf{pobj}] he came[\textbf{relcl}]  from.&A reef encloses the cove&he comes from the cove.\\
		\hline
		Relative-nsubj&Frank gestures to the SALESMAN, \emph{who}[\textbf{nsubj}]'s waiting[\textbf{relcl}]  on a woman&the SALESMAN waits on a woman.&Frank gestures to the SALESMAN.\\
		\hline
		Relative-advmod&Chuck is in the stage of exposure \emph{where}[\textbf{advmod}] the personality splits[\textbf{relcl}] &Chuck is in the stage of exposure&the personality splits at exposure.\\
		\hline
		Relative-poss&The girl, \emph{whose}[\textbf{poss}] name is[\textbf{relcl}] Helga, cowers.&The girl cowers&The girl 's name is Helga\\
		\hline
		Relative-omit&Kim is the sexpot Peter saw[\textbf{relcl}] in Washington Square Park&Peter sees Kim in Washington Square Park.&Kim is the sexpot.\\
		\hline
		Adverbial&Jim panics as[\textbf{advcl}] his mom reacts, shocked.&Jim panics, shocked.& Jim's mom reacts.\\
		\hline
		Adverbial-remove&Suddenly there's a KNOCK at the door, immediately after[\textbf{prep}] which JIM'S MOM enters[\textbf{advcl}].&Suddenly there 's a KNOCK at the door.& Immediately JIM 'S MOM enters.\\
		\hline
		Inverted Cl. Subject&Running[\textbf{csubj}] towards Oz is Steve Stifler&Steve Stifler runs towards Oz.& --\\
		\hline
		Clausal Component&The thing is, it actually sounds[\textbf{ccomp}] really good.& The thing is.(will be eliminated by the filter)&It actually sounds really good.\\
		\hline
		Passive Voice&They[\textbf{nsubjpass}] are suddenly illuminated by the glare of headlights.& Suddenly the glare of headlights illuminateds them.&--\\
		\hline
		Open Clausal&The sophomore comes running[\textbf{xcomp}] through the kitchen.& The sophomore runs through the kitchen.& The sophomore comes. \\
		\hline
		Adjective& Stifler has a toothbrush hanging[\textbf{acl}] from his mouth.&A toothbrush hangs from Stifler's mouth.& Stifler has a toothbrush.\\
		\hline
	\end{tabular}
    \caption[Results of Simplification]{Syntactic simplification rules applied on example sentences. \label{tab:simp-res}}
\end{table*}


There has been work on text simplification using linguistic rules-based approaches. For example, \cite{Siddharthan:2011} propose a set of rules to manipulate sentence structure to output simplified sentences using syntactic dependency parsing. Similarly, the YATS system \cite{YATS} implements a set of rules in the JAPE language \cite{JAPE} to address six syntactic structures: \textbf{Passive Constructions, Appositive Phrases, Relative Clauses, Coordinated Clauses, Correlated Clauses} and \textbf{Adverbial Clauses}. Most of the rules focus on case and tense correction, with only 1-2 rules for sentence splitting. 
We take inspiration from the YATS system, and our system incorporates modules to identify and transform sentence structures into simpler ones using a broader set of rules.  

In our system, each syntactic structure is handled by an \textit{Analyzer}, which contains two processes: \textit{Identify} and \textit{Transform}. The \textit{Identify} process takes in a sentence and determines if it contains a particular syntactic structure. Subsequently, the \textit{Transform} process focuses on the first occurrence of the identified syntactic structure and then splits and assembles the sentence into simpler sentences. Both \textit{Identify} and \textit{Transform} use Part-of-Speech (POS) tagging and dependency parsing \cite{spacy2} modules implemented in 
spaCy 2.0\footnote{\url{https://spacy.io}}

The simplification algorithm (Algorithm~\ref{alg:syntac-sys}) starts with an input sentence  and recursively processes it until no further simplification is possible. It uses a queue to manage intermediate simplified sentences, and runs in a loop until the queue is empty. For each sentence, the system applies each syntactic analyzer to \textit{Identify} the corresponding syntactic structure in the sentence (line 14). If the result is positive, the sentence is processed by the \textit{Transform} function to convert it to simple sentences (line 16). Each of the output sentences is pushed by the controller into the queue (line 19). The process is repeated with each of the \textit{Identify} analyzers (line 13). If none of the analyzers can be applied, the sentence is assumed to be simple and it is pushed into the result list (line 21). We summarize linguistic rules in Table \ref{tab:simp-rules} and examples are given in Table \ref{tab:simp-res}. Next, we describe the coordination linguistic rules. For details regarding other rules, please refer to Appendix B. 

\noindent\textbf{Coordination:} \label{coor}
Coordination is used for entities having the same syntactic relation with the head and serving the same functional role (e.g. subj, obj, etc.). 
It is the most important component in our simplification system. The parser tags word units such as 
``and'' and ``as well as'' with the dependency label \textit{cc}, and the conjugated words as \textit{conj}. Our system deals with coordination based on the dependency tag of the conjugated word. 

\begin{figure*}[t]
	\centering
\includegraphics[width=\linewidth]{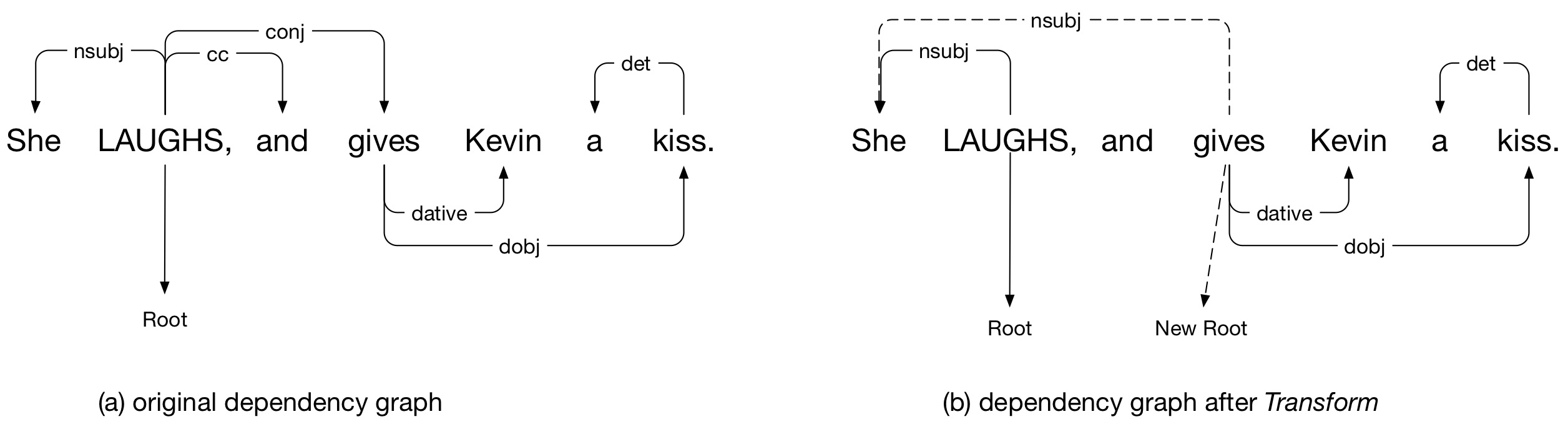}\\
\caption[Coordination demo]{\emph{Transform} in an example coordination sentence. Firstly the dependency links of \emph{cc} and \emph{conj} are cut. Then we look for a noun in the left direct children of the original root \emph{LAUGHS} and link the new root \emph{gives} with it. In-order traverse from the original root and the new root will result in simplified sentences as shown in Table~\ref{tab:simp-res}.
		\label{fig:indent-dist}}
\end{figure*}

In the case of coordination, the \emph{Identify} function simply returns whether \emph{cc} or \emph{conj} is in the dependency graph of the input sentence. The \emph{Transform} function 
manipulates the graph structure based on the dependency tags of the conjugated words as shown in Figure~\ref{fig:indent-dist}. If the conjugated word is a verb, then we mark it as another root of the sentence. Cutting \emph{cc} and \emph{conj} edges in the graph and traversing from this new root results in a new sentence parallel to the original one. In other cases, such as the conjugation between nouns, we simply replace the noun phrases with their siblings and traverse from root again. 

\subsection{Lexical Simplification}
In order to generate animation, actions and participants extracted from simplified sentences are mapped to existing actions and objects in the animation engine. Due to practical reasons, it is not possible to create a library of animations for all possible actions in the world.  We limit our library to a predefined list of 52 actions/animations, expanded to 92 by a dictionary of synonyms (\textsection \ref{sec:animation}). We also have a small library of pre-uploaded objects (such as ``campfire'', ``truck''  and others).  

To animate unseen actions not in our list, we use a word2vec-based \textit{similarity} function to find the nearest action in the list. Moreover, we use WordNet \cite{wordnet} to exclude antonyms. This helps to map non-list actions (such as ``squint at'') to the similar action in the list (e.g. ``look''). If we fail to find a match, we check for a mapping while including the verb's preposition or syntactic object. We also use WordNet to obtain hypernyms for further checks, 
when the \emph{similarity} function fails to find a close-enough animation. 
 Correspondingly, for objects, we use the same \emph{similarity} function and  WordNet's holonyms.
 
 Our list of actions and objects is not exhaustive. Currently, we do not cover actions which may not be visual. For out of list actions, we give the user a warning that the action cannot be animated. Nevertheless, this is a work in progress and we are working on including more animations for actions and objects in our knowledge base. 
 


\subsection{Action Representation Field (ARF): Information Extraction}
\label{sec:srl}

For each of the simplified sentences, information is extracted and populated into a predefined key-value pair structure. We will refer to the keys of this structure as \textit{Action Representation Fields (ARFs)}. These are similar to entities and relations in Knowledge Bases. ARFs include: \textit{owner}, \textit{target}, \textit{prop}, \textit{action}, \textit{origin\_action}, \textit{manner}, \textit{modifier\_location}, \textit{modifier\_direction}, \textit{start-time}, \textit{duration}, \textit{speed}, \textit{translation}, \textit{rotation}, \textit{emotion}, \textit{partial\_start\_time} (for more details see Appendix C). This structure is inspired by the PAR \cite{PAR} architecture, but adapted to our needs. 


To extract the ARFs from the simplified sentences, we use a Semantic Role Labelling (SRL) model in combination with some heuristics, for example creating word lists for \emph{duration, speed, translation, rotation, emotion}. We use a pre-trained Semantic Role Labelling model\footnote{AllenNLP SRL model: \url{https://github.com/allenai/allennlp}} based on a Bi-directional LSTM network \cite{he2017deep} with pre-trained ELMo embeddings~\cite{peters2018deep}. We map information from each sentence to the knowledge base of animations and objects.

\subsection{Animation  Generation}
\label{sec:animation}
We use the animation pipeline of the CARDINAL system. 
We plug in our NLP module in CARDINAL to generate animation. CARDINAL creates pre-visualizations of the text, both in storyboard form and animation. A storyboard is a series of pictures that demonstrates the sequence of scenes from a script. The animation is a 3-D animated video that approximately depicts the script. CARDINAL uses the Unreal game engine \cite{games2007unreal} for generating pre-visualizations. It has a knowledge base of pre-baked animations (52 animations, plus a dictionary of synonyms, resulting in 92) and pre-uploaded objects (e.g. “campfire”, “tent”). It also has 3-D models which can be used to create characters. 

%% file: data.tex

\begin{table*}[t]
    \footnotesize
	\centering
	\begin{tabular}{|p{0.25\textwidth}||p{0.25\textwidth}|p{0.25\textwidth}|p{0.15\textwidth}|}
		\hline
		\multicolumn{4}{|c|}{\textbf{Carl touches Ellie's shoulder as the doctor explains. Ellie drops her head in her hands.}}\\
		\hline
		System output & Annotator \rom{1}&Annotator \rom{2}&$BLEU_{2}$(\%)\\
		\hline
		Carl touches Ellie 's shoulder
		& carl touches ellie's shoulder&
		carl touches ellie's shoulder.
		&38.73\\
		\hline
		the doctor explains& 
		the doctor explains&
		the doctor is talking.&100\\
		\hline
		Ellie drops Ellie head in Ellie hands&
		ellie drops her head in her hands&
		ellie drops her head in her hands.
		&48.79\\
		\hline
	\end{tabular}
	\caption[Differences between System output and Ground Reference]{Differences between system output and annotator responses \label{tab:error-cases}}
\end{table*}

\begin{table}[t]
    \small
	\centering
	\begin{tabular}{|c|c|c|}
		\hline
		&$BLEU$ & $SARI$ \\
		\hline
		NTS-w2v & 61.45 & 36.04\\
		\hline
		 YATS & 58.83 & 48.75\\
		\hline
		Our System & \textbf{67.68} & \textbf{50.65}\\
		\hline
	\end{tabular}
	\caption[Results on syntactic simplification]{Results on syntactic simplification \label{tab:bleu}}
\end{table}

\section{Text-to-Animation Corpus} \label{sec:data}

We initially used a corpus of Descriptions components from ScreenPy \cite{Screenpy}, in order to study linguistic patterns in the movie script domain. Specifically, we used the ``heading'' and ``transition'' fields from ScreenPy's published JSON output on 1068 movie scripts scraped from IMSDb.
We also scraped screenplays from SimplyScripts and ScriptORama\footnote{\url{http://www.simplyscripts.com} and \url{http://www.script-o-rama.com}}. After separating screenplays into well-formatted and ill-formatted, Descriptions components were  extracted using our model (\textsection \ref{sec:parsing}). This gave a corpus of Descriptions blocks from 996 screenplays.

The corpus contains a total of 525,708 Descriptions components. The Descriptions components contain a total of 1,402,864 sentences. Out of all the Descriptions components, 49.45\% (259,973) contain at least one verb which is in the animation list (henceforth called ``action verbs''). Descriptions components having at least one action verb have in total 920,817 sentences. Out of these, 42.2\% (388,597) of the sentences contain action verbs. 
In the corpus, the average length of a sentence is 12 words.

%% file: evaluation.tex

\section{Evaluation and Analysis} \label{sec:evaluation}

There are no standard corpora for text-to-animation generation. It is also not clear how should such systems be evaluated and what should be the most appropriate evaluation metric. Nevertheless, it is important to assess how our system is performing. We evaluate our system using two types of evaluation: Intrinsic and Extrinsic. Intrinsic evaluation is for evaluating the NLP pipeline of our system using the BLEU metric. Extrinsic evaluation is an end-to-end qualitative evaluation of our text-to-animation generation system, done via a user study. 
\subsection{Intrinsic Evaluation} \label{sec:intrinsic}
To evaluate the performance of our proposed NLP pipeline, 500 Descriptions components from the test set were randomly selected. Three annotators manually translated these 500 Descriptions components into simplified sentences and extracted all the necessary ARFs from the simplified sentences. This is a time intensive process and took around two months. 30\% of the Descriptions blocks contain verbs not in the list of 92 animation verbs. There are approximately 1000 sentences in the test set, with average length of 12 words. Each Descriptions component is also annotated by the three annotators for the ARFs. 


Taking inspiration from the text simplification community \cite{nisioi2017exploring,saggion2017automatic}, we use the BLEU score \cite{BLEU} for evaluating our simplification and information extraction modules.  For each simplified sentence $s_i$ we have 3 corresponding references $r^1_i, r^2_i$ and $r^3_i$. 
We also evaluate using the SARI \cite{xu2016optimizing} score to evaluate our text simplification module. 


\begin{table*}[t]
\begin{tabular}{cc}
\noindent
\hspace{2pt}
\begin{minipage}{.45\textwidth}
    \small
	\centering
	\begin{tabular}[H]{|l|l|l|l|}
		\hline
		Field&$BLEU_{1}$&Field&$BLEU_1$\\
		\hline
		\textbf{owner}&56.16&\textbf{org\_action}&80.92\\
		\textbf{target}&41.85&\textbf{manner}&84.84\\
		\textbf{prop}&28.98&\textbf{location}&71.89\\
		\textbf{action}&70.46&\textbf{direction}&70.83\\
		\textbf{emotion}&57.89&&\\
		\hline
	\end{tabular}
	\caption[Results on textual ARFs]{Results on textual ARFs (\%)
	\label{tab:text-fields}}

\end{minipage}%
&\hspace{20pt}
\begin{minipage}{.45\textwidth}
    \small
	\centering
	\begin{tabular}[H]{|l|l|l|l|} 
		\hline
		Field&\emph{P}&\emph{R}&\emph{F1}\\
		\hline
		\textbf{s\_time}&86.49&68.63&76.53\\
		\textbf{rot.}&82.04&81.16&81.60\\
		\textbf{duration}&94.72&73.92&83.04\\
		\textbf{transl.}&75.49&86.47&80.61\\
		\textbf{speed}&94.41&79.50&86.32\\
		\hline
	\end{tabular}
	\caption[Result on Non-textual ARFs]{Result on Non-textual ARFs(\%) \label{tab:none-text-fields}}
\end{minipage}
\end{tabular}
\end{table*}

\subsubsection{Sentence Simplification} 

Each action block $a$ is reduced to a set of simple sentences $S_a=\{s_1, s_2,....s_{n_a}\}$. And for the same action block $a$, each annotator $t$, $t\in \{1,2,3\}$ produces a set of simplified sentences $R^t_a=\{r^t_1,r^t_2,...r^t_{m^t_a}\}$. Since the simplification rules in our system may not maintain the original ordering of verbs, we do not have sentence level alignment between elements in $S_a$ and $R^t_a$. For example, action block $a$ = \textit{He laughs after he jumps into the water} is reduced by our system  into two simplified sentences $S_a=\{s_1=\textit{He jumps into the water}, s_2=\textit{He laughs}\}$ by the temporal heuristics, while $annotator_3$ gives us $R^3_a=\{r^3_1=\textit{He laughs}, r^3_2=\textit{He jumps into the water}\}$. In such cases, sequentially matching $s_i$ to $r_j$ will result in a wrong (hypothesis, reference) alignment which is ($s_1$, $r^3_1$) and ($s_2$, $r^3_2$).

To address this problem, for each hypothesis $s_i\in S_a$, we take the corresponding reference $r^t_i\in R^t_a$ as the one with the least Levenshtein Distance \cite{navarro2001guided} to $s_i$, i.e, $$r^t_i=\argmin_{r^t_j} lev\_dist(s_i, r^t_j), \forall j\in\{1,...,m^t_a\}$$

As per this alignment, in the above example, we will have correct alignments ($s_1$, $r^3_2$) and ($s_2$, $r^3_1$). Thus, for each simplified sentence $s_i$ we have 3 corresponding references $r^1_i$, $r^2_i$ and $r^3_i$. The aligned sentences are used to calculate corpus level BLEU score\footnote{We used NLTK's API with default parameters: \url{http://www.nltk.org/api/nltk.translate.html\#nltk.translate.bleu_score.corpus_bleu}} and SARI score\footnote{Implementation available at \url{https://github.com/cocoxu/simplification/}}. 




The evaluation results for text simplification are summarized in Table~\ref{tab:bleu}. We compare against YATS \cite{YATS} and neural end-to-end text simplification system NTS-w2v \cite{nisioi2017exploring}. YATS is also a rule-based text simplification system. As shown in Table \ref{tab:bleu}, our system performs better than YATS on both the metrics, indicative of the limitations of the YATS system. A manual examination of the results also showed the same trend. However, the key point to note is that we are not aiming for text simplification in the conventional sense. Existing text simplification systems tend to summarize text and discard some of the information. Our aim is to break a complex sentence into simpler ones while preserving the information. 

An example of a Descriptions component with $BLEU_{2}$ scores is given in Table~\ref{tab:error-cases}. In the first simplified sentence, the space between \emph{Ellie} and \emph{'s} causes the drop in the score. But it gives exactly the same answer as both annotators. In the second sentence, the system output is the same as the annotator \rom{1}'s answer, so the $BLEU_{2}$ score is 1. In the last case, the score is low, as annotators possibly failed to replace \emph{her} with the actual Character Cue \emph{Ellie}. Qualitative examination reveals, in general, that our system gives a reasonable result for the syntactic simplification module. As exemplified, BLEU is not the perfect metric to evaluate our system, and therefore in the future we plan to explore other metrics. 

\subsubsection{ARF Evaluation}
We also evaluate the system's output for action representation fields against gold annotations. In our case, some of the fields can have multiple (2 or 3) words such as \emph{owner, target, prop, action, origin\_action, manner, location} and \emph{direction}. We use $BLEU_{1}$ as the evaluation metric to measure the BOW similarity between system output and ground truth references. The results are shown in Table~\ref{tab:text-fields}.

In identifying \emph{owner, target} and \emph{prop}, the system tends to use a fixed long mention, while annotators prefer short mentions for the same character/object. The score of \emph{prop} is relatively lower than all other fields, which is caused by a systematic SRL mapping error. The relatively high accuracy on the \emph{action} field indicates the consistency between system output and annotator answers. 

Annotation on the \emph{emotion} ARF is rather subjective. Responses on the this field are biased and noisy. The $BLEU_{1}$ score on this is relatively low. 
For the other non-textual ARFs, we use precision and recall to measure the system's behavior. Results are shown in Table~\ref{tab:none-text-fields}. These  fields are subjective: annotators tend to give different responses for the same input sentence.

\emph{rotation} and \emph{translation} have Boolean values. Annotators agree on these two fields in most of the sentences. The system, on the other hand, fails to identify actions involving \emph{rotation}. For example, in the sentence \emph{``Carl closes CARL 's door sharply''} all four annotators think that this sentence involves rotation, which is not found by the system. This is due to the specificity of rules on identifying these two fields. 

\emph{speed}, \emph{duration} and  \emph{start\_time} have high precision and low recall. This indicates the inconsistency in annotators' answers. For example, in the sentence \emph{``Woody runs around to the back of the pizza truck''}, two annotators give 2 seconds and another gives 1 second in \emph{duration}. These fields are subjective and need the opinion of the script author or the director. In the future, we plan to involve script editors in the evaluation process. 

\begin{table*}[t]
\footnotesize
\centering
\begin{tabular}{|p{0.37\linewidth}|p{0.1\linewidth}|l|l|l|p{0.1\linewidth}|}
\hline
\small
\textbf{}                                                                   & Strongly Disagree & Disagree & Neutral & Agree & Strongly Agree \\ \hline
The animation shown in the video is a reasonable visualization of the text. & 13.64 \%                      & 18.18 \%          & 22.95 \%         & 28.64 \%       & 16.59 \%                \\ \hline
All the actions mentioned in the text are shown in the video.               & 15.68 \%                   & 20 \%          & 22.96 \%         & 20.68 \%       & 20.68 \%                \\ \hline
All the actions shown in the video are mentioned in the text.               & 12.96 \%                   & 11.14 \%          & 16.36 \%         & 23.18 \%       & 36.36 \%                \\ \hline
\end{tabular}
\caption[User Study Results]{User Study Results \label{tab:user-study}}
\end{table*}

\subsection{Extrinsic Evaluation} \label{qualitative}
We conducted a user study to evaluate the performance of the system qualitatively. The focus of the study was to evaluate (from the end user's perspective) the performance of the NLP component w.r.t. generating reasonable animations. 

We developed a questionnaire consisting of 20 sentence-animation video pairs. The animations were generated by our system. 
The questionnaire was filled by 22 participants. On an average it took around 25 minutes for a user to complete the study.  

We asked users to evaluate, on a five-point \emph{Likert scale} \cite{likert1932}, if the video shown was a reasonable animation for the text, how much of the text information was depicted in the video and how much of the information in the video was present in the text (Table~\ref{tab:user-study}). The 68.18\% of the participants rated the overall pre-visualization as neutral or above. The rating was 64.32\% (neutral or above) for the conservation of textual information in the video, which is reasonable, given limitations of the system that are not related to the NLP component. For the last question, 75.90\% (neutral or above) agreed that the video did not have extra information. 
In general, there seemed to be reasonable consensus in the responses. Besides the limitations of our system, disagreement can be attributed to the ambiguity and subjectivity of the task.

We also asked the participants to describe qualitatively what textual information, if any, was missing from the videos. Most of the missing information was due to limitations of the overall system rather than the NLP component: facial expression information was not depicted because the character 3-D models are deliberately designed without faces, so that animators can draw on them. Information was also missing in the videos if it referred to objects or actions that do not have a close enough match in the object list or animations list. Furthermore, the animation component only supports animations referring to a character or object as a whole, not parts, (e.g. ``Ben raises his head'' is not supported).

However, there were some cases where the NLP component can be improved. For example, lexical simplification failed to map the verb ``watches''  to the similar animation ``look''. In one case, syntactic simplification created only two simplified sentences for a verb which had three subjects in the original sentence. In a few cases, lexical simplification successfully mapped to the most similar animation (e.g.``argue'' to ``talk'') but the participants were not satisfied - they were expecting a more exact animation. We plan to address these shortcomings in future work. 


%% file: conclusion.tex
\section{Conclusion and Future Work} \label{sec:conclusion} \vspace{-1mm}
In this paper, we proposed a new 
text-to-animation system. The system uses linguistic text simplification techniques to map screenplay text to animation. Evaluating such systems is a challenge. Nevertheless, intrinsic and extrinsic evaluations show reasonable performance of the system. The proposed system is not perfect, for example,  the current system does not take into account the discourse information that links the actions implied in the text, as currently the system only processes sentences independently. In the future, we would like to leverage discourse information by considering the sequence of actions which are described in the text \cite{modi2014inducing,modi2016event}. This would also help to resolve ambiguity in text with regard to actions \cite{TACL968,Modi17}. Moreover, our system can be used for generating training data which could be used for training an end-to-end neural system. 

\section*{Acknowledgments}

We would like to thank anonymous reviewers for their insightful comments. We would
also like to thank Daniel Inversini, Isabel Simó Aynés, Carolina Ferrari, Roberto Comella and Max Grosse for their help and support in developing the Cardinal system. Mubbasir Kapadia has been funded in part by NSF IIS-1703883 and NSF S\&AS-1723869.

%% file: Appendix/appendixa.tex
\section*{Appendix A}\label{appendix:0}

{\centering\begin{figure}[!h]\onecolumn
        \center{\includegraphics[width=0.5\textwidth]{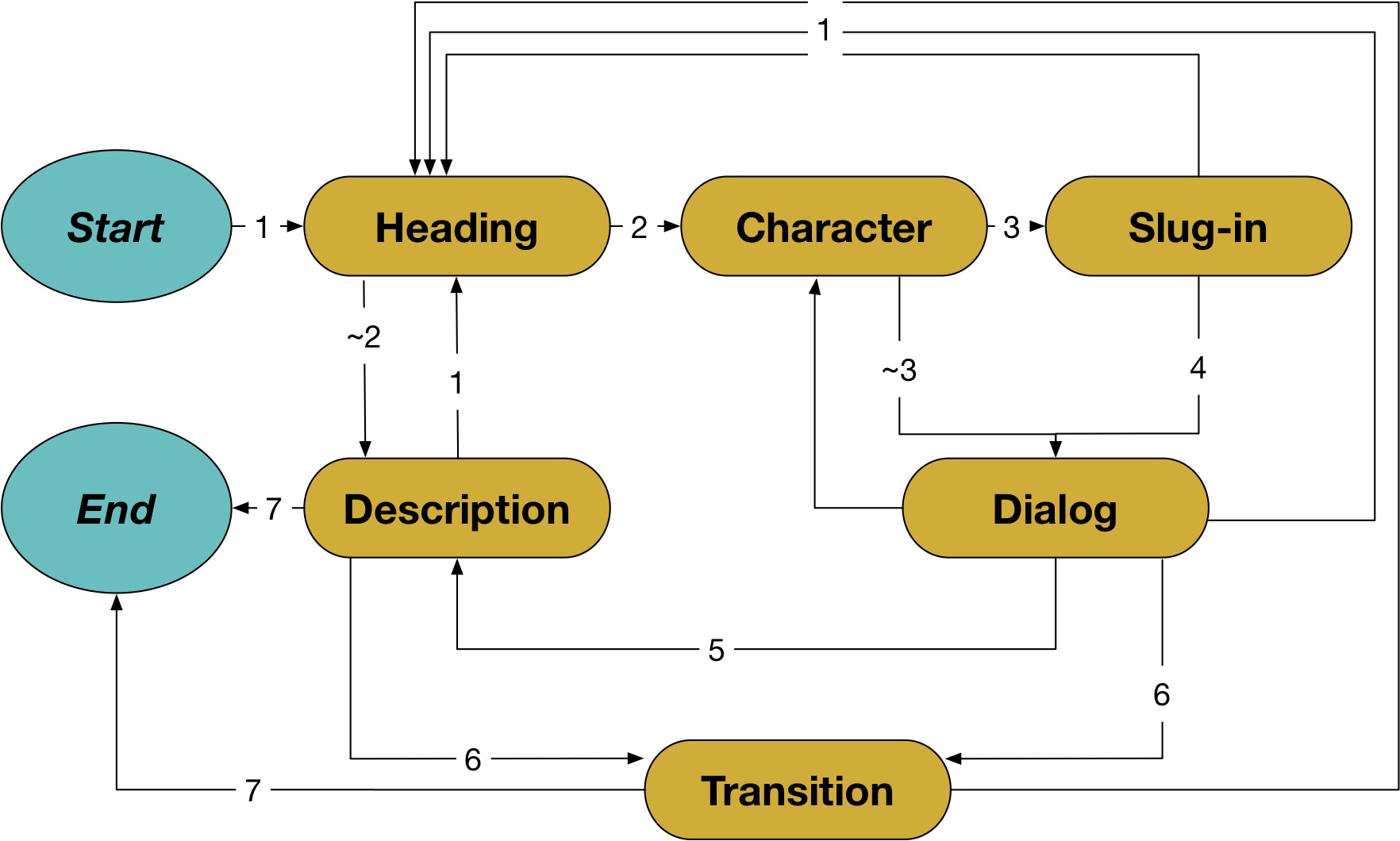}
        \caption{\small FSM of Well-formatted Screenplay Parser. Numbers of link are Rules and $\sim$ means logical NOT}
        \label{fig:wellparsefsm}}
\end{figure}
\vspace{10mm}

\begin{table}[!h]
 \centering
\begin{tabular}{|p{0.025\textwidth}|p{0.95\textwidth}|}
		\hline
		ID & Rule Summary\\
		\hline
		1 & If input is uppercase and contains heading words such as `INT', `EXT'. etc, return True, otherwise False \\
		\hline
		2 &  If input is uppercase and contains character words such as `CONT.', `(O.S)', or if \#indentation $>$ most frequent \#indentation, return True. Otherwise False\\
		\hline
		3& If input starts with \(`('\) , return True, otherwise False\\
		\hline
		4& If input ends with \(`)'\), return True, otherwise False\\
		\hline
		5& If $\abs{\#last indents-\#current indents} < 3$, return True, otherwise False \\
		\hline
		6& If the input is uppercase and contains transition words such as `DISSOLVE', `CUT TO'. etc, return True, otherwise False\\
		\hline
		7& If the input equals to `THE END', return True. Otherwise False.\\
		\hline
\end{tabular}
	\caption{FSM Transition Rules} 
	\label{tab:parse-rule-all}
\end{table}


\begin{table*}[!htb]
	\centering
	\begin{tabular}{|p{\textwidth}|}
		\hline
		COMPLEX: Another parent , Mike Munson , sits on the bench with a tablet and uses an app to track and analyze the team ’s shots.\\
		\hline
		NSELSTM-B: Another parent, Mike Munson, sits on the bench with a tablet and uses an app to track. \\
		\hline
		YATS: Another parent sits on the bench with a tablet and uses an app to track and examines the team’ s shots. This parent is Mike Munson. \\
		\hline
		OURS: Another parent is Mike Munson. Another parent sits on the bench with a tablet. Another parent uses an app.\\
		\hline\hline
		COMPLEX: Stowell believes that even documents about Lincoln’s death will give people a better understanding of the man who was assassinated 150 years ago this April. \\
		\hline
		NSELSTM-B: Stowell believes that the discovery about Lincoln’s death will give people a better understanding of the man. \\
		\hline
		YATS: Stowell believes that even documents about Lincoln’ s death will give people a better reason of the man. This man was assassinated 150 years ago this April. \\
		\hline
		OURS: Stowell believes. Even documents about Lincoln ’s death give people a better understanding of the man. Somebody assassinates the man 150 years ago this April.\\
		\hline
		\end{tabular}
	\caption{Example Model Outputs}
	\label{tab:simp-result-example}
\end{table*}
}

%% file: Appendix/appendixb.tex
\section*{Appendix B: Algorithms}\label{appendix:A}

\begin{algorithm}[!hb]
    \scriptsize
	\caption{Identify Adverbial Clause}\label{alg:adverbial-identify}
	\begin{algorithmic}[1]
		\Procedure{Adverbial Clause identify Procedure}{$sent$}\Comment{The input sentence}
		\State find tokens in sents with dependency tag ROOT and ADVCL (we call it advcl)
		\If{no ADVCL token in sents}
		\State \textbf{return} False
		\EndIf
		\State find father token of \textit{advcl} as  \textit{father}
		\If{\textit{father} is not a VERB}
		\State we make it as a VERB \Comment{We correct POS error here}
		\EndIf
		\State find conjunction tokens of \textit{father} as \textit{conjuncts}
		\If{NOUN in \textit{advcl}'s left subtree}
		\State subject $\gets$ this NOUN
		\Else
		\State subject $\gets$ NOUN in \textit{father}'s left subtree
		\EndIf
		\State \textbf{return} True
		\EndProcedure
	\end{algorithmic}
\end{algorithm}

\begin{algorithm}[!htb]
    \scriptsize
	\caption{Transform Adverbial Clause}\label{alg:adverbial-transform}
	\begin{algorithmic}[1]
		\Procedure{Adverbial Clause transform Procedure}{$sent$}\Comment{The input sentence}
		\State cut edge between root and advcl token\Comment{remove advcl token from root's children}
		\If{advcl verb does not have its own subject}
		\State add subject as advcl's most left direct child
		\EndIf
		\State remove PREP and MARK token in advcl's children, modify temporal id accordingly
		\State str1 $\gets$ traverse\_a\_string(root)
		\State str2 $\gets$ correct\_tense(traverse\_a\_string(advcl))
		\State \textbf{return} [str1, str2]
		\EndProcedure
	\end{algorithmic}
\end{algorithm}


\begin{algorithm}[!htb]
    \scriptsize
	\caption{Identify() in Relative Clause Analyzer}\label{alg:recl-identify}
	\begin{algorithmic}[1]
		\Procedure{Relative Clause identify Procedure}{$sent$}\Comment{The input sentence.}
		\If{no RELCL token in sent}
		\State  \textbf{return} False
		\EndIf
		\State anchor $\gets$ RELCL token
		\State head $\gets$ anchor's head token
		\State wg $\gets$ NULL
		\For{token t in anchor's children}
		\If{t.tag $\in$ [WDT, WP, WP\$, WRB]}
		\State wh $\gets$ t
		\EndIf
		\EndFor
		\State \textbf{return} True
		\EndProcedure
	\end{algorithmic}
\end{algorithm}

\begin{algorithm}[!htb]
    \scriptsize
	\caption{Transform() in Relative Clause Analyzer}\label{alg:recl-transform}
	\begin{algorithmic}[1]
		\Procedure{Relative Clause transform Procedure}{$root, anchor, head, wh$}\Comment{Root of dependency tree, the recl anchor token, its head, and wh-word}
		\State cut relcl edge between head and anchor
		\State str1 $\gets$ traverse\_a\_string(root)
		\If{wh $=$ NULL}\Comment{No Wh-word in the sentence, concatenate}
		\State str2 $\gets$traverse\_a\_string(anchor) + traverse\_a\_string(head)
		\Else
		\State wh-dep $\gets$ dependency tag of wh
		\State wh-head $\gets$ head of wh
		\State remove wh in anchor's children
		\If{wh-dep$=$ DOBJ}\Comment{wh is verb}
		\State str2$\gets$ traverse\_a\_string(wh-head) + traverse\_a\_string(anchor) 
		\ElsIf{wh-dep$=$ POBJ}\Comment{wh-head is preposition}
		\State put head after wh-head in anchor's children
		\State str2 $\gets$ traverse\_a\_string(anchor) 
		\ElsIf{wh-dep$\in$ [NSUBJ, NSUBJPASS]}\Comment{wh is subject}
		\State str2 $\gets$ traverse\_a\_string(wh-head) + traverse\_a\_string(anchor)
		\ElsIf{wh-dep$=$ ADVMOD}\Comment{wh is time or location}
		\State prep $\gets$ `at'
		\State str2 $\gets$ traverse\_a\_string(anchor) + prep+ traverse\_a\_string(wh-head)
		\ElsIf{wh-dep$=$ POSS}\Comment{wh $=$ whose}
		\State str2 $\gets$ traverse\_a\_string(wh-head) + be verb+ traverse\_a\_string(anchor)
		\EndIf 
		\EndIf
		\State correct verb tense in str1 and str2
		\State \textbf{return} [str1, str2]
		\EndProcedure
	\end{algorithmic}
\end{algorithm}


\begin{algorithm}[!htb]
    \scriptsize
	\caption{Transform() in Coordination Analyzer}\label{alg:coord-transform}
	\begin{algorithmic}[1]
		\Procedure{Coordination transform Procedure}{$sents$}\Comment{Input sentence}
		\State results $\gets$ empty list
		\State find \emph{root} of dependency tree of \emph{sents}
		\State find first \emph{cc}(if any) and \emph{conj} token in denpendency tree and their head token \emph{main}
		\State embed all conjugate words of \emph{main} in a list \emph{conjus}
		\State cut conj edge between \emph{main} and \emph{cc} and \emph{conj}
		\If{no object for \emph{main}}
		\State try find object in \emph{conj}'s right children
		\EndIf
		\State str1 $\gets$ traverse\_a\_string(\emph{root})
		\State results.append(str1)
		\For{\emph{conj} in \emph{conjs}}
		\State \emph{type} $\gets$ get\_conj\_type(\emph{main}, \emph{conj})
		\If{\emph{type}$=$VERB\&VERB}
		\State correct part-of-speech tag if necessary\Comment{spaCy tends to tag verb as noun}
		\If{\emph{conj} has its own subject}
		\State \emph{new-root} $\gets$ \emph{conj}
		\Else
		\If{\emph{main}$=$\emph{root}}
		\State \emph{new-root} $\gets$ \emph{conj}
		\Else
		\State replace \emph{main} with \emph{conj} in \emph{root}'s children
		\State \emph{new-root} $\gets$ \emph{root}
		\EndIf
		\EndIf
		\Else \Comment{Other cases such as NOUN\&NOUN, AD*\&AD*, apply same rule}
		\State \emph{main-head} $\gets$ head of \emph{main}
		\State replace \emph{main} with \emph{conj} in \emph{main-head}'s children
		\State \emph{new-root} $\gets$ \emph{root}
		\EndIf
		\State str2 $\gets$ traverse\_a\_string(\emph{new-root})
		\State results.append(str2)
		\EndFor
		\State \textbf{return} results
		\EndProcedure
	\end{algorithmic}
\end{algorithm}


\begin{algorithm}[!htb]
    \scriptsize
	\caption{Identify() in Passive Analyzer}\label{alg:passive-identify}
	\begin{algorithmic}[1]
		\Procedure{Passive Voice identify Procedure}{$sents$}\Comment{Input sentence}
		\State is-passive $\gets$ False
		\For{token \emph{t} in sents}
		\If{t.dep $\in$ [CSUBJPASS, NSUBJPASS]}
		\State subj-token $\gets$ t
		\State verb-token $\gets$ t.head
		\State is-passive $\gets$ True
		\EndIf
		\If{t.dep $\in$  [AUXPASS] and t.head $=$ verb-token}
		\State auxpass-token $\gets$ t
		\EndIf
		\If{t.dep $\in$ [AGENT] and t.text $=$ `by'}
		\State by-token $\gets$ t
		\EndIf
		\EndFor
		\State \textbf{return} is-passive
		\EndProcedure
	\end{algorithmic}
\end{algorithm}
\begin{algorithm}[!htb]
    \scriptsize
	\caption{Transform() in Passive Analyzer}\label{alg:passive-transform}
	\begin{algorithmic}[1]
		\Procedure{Passive Voice transform Procedure}{$sents$}\Comment{Input sentence}
		\If{auxpass-token $\ne$ NULL}
		\State cut auxpass edge
		\EndIf
		\State cut nsubjpass or csubj edge
		\State prepend subject-token to verb-token's right children
		\If{by-token $\ne$ NULL}
		\State cut by-agent edge
		\State add by-token's right children to verb-token's left children
		\Else
		\State add `Somebody' to verb-token's left children
		\EndIf
		\State correct verb tense for verb-token
		\State \textbf{return} traverse\_a\_string(root-token)
		\EndProcedure
	\end{algorithmic}
\end{algorithm}


\begin{algorithm}[!htb]
    \scriptsize
	\caption{Transform() in Appositive Clause Analyzer}\label{alg:appos-transform}
	\begin{algorithmic}[1]
		\Procedure{Appositive Clause transform Procedure}{$anchor, head$}\Comment{The APPOS token and its head token.}
		\State cut edge between \textbf{anchor} and \textbf{head} token\Comment{remove \textbf{anchor} from \textbf{head}'s right children}
		\State str1 $\gets$ traverse\_a\_string(root token of input sentence)
		\State str2 $\gets$ traverse\_a\_string(\textbf{head}) $+$ be\_verb $+$ traverse\_a\_string(\textbf{anchor})
		\State correct verb tense in str1 and str2
		\State \textbf{return} [str1, str2]
		\EndProcedure
	\end{algorithmic}
\end{algorithm}

\begin{algorithm}[!htb]
    \scriptsize
	\caption{Identify() in Inverted Clausal Subject Analyzer}\label{alg:subj-identify}
	\begin{algorithmic}[1]
		\Procedure{Inverted Clausal Subject identify Procedure}{$sents$}\Comment{Input sentence}
		\For{Token t in \emph{sents}}
		\If{t.dep $=$ CSUBJ and t.tag $\in$ [VBN, VBG] and t.head.lemma$=$`be'}
		\State attr $\gets$ token with dependency label \emph{attr} in t.head's right children
		\If{attr $=$ NULL} \Comment{\emph{attr} is the actual subject of the sentence}
		\State \textbf{return} False \Comment{Make sure this is an inverted sentence}
		\EndIf
		\State actual-verb $\gets$ t
		\State be-verb $\gets$ t.head
		\State \textbf{return} True
		\EndIf
		\EndFor
		\State \textbf{return} False
		\EndProcedure
	\end{algorithmic}
\end{algorithm}

\begin{algorithm}[!htb]
    \scriptsize
	\caption{Transform() in Inverted Clausal Subject Analyzer}\label{alg:subj-transform}
	\begin{algorithmic}[1]
		\Procedure{Inverted Clausal Subject transform Procedure}{$sents$}\Comment{Input sentence}
		\State get access to actual-verb, be-verb and attr in identify procedure~\ref{alg:subj-identify}
		\State change position of actual-verb and attr in be-verb's children
		\State \textbf{return} [traverse\_a\_string(be-verb)]
		\EndProcedure
	\end{algorithmic}
\end{algorithm}

\begin{algorithm}[!htb]
    \scriptsize
	\caption{Transform() in CCOMP Analyzer}\label{alg:ccomp-transform}
	\begin{algorithmic}[1]
		\Procedure{Clause Component transform Procedure}{$sents$}\Comment{Input sentence}
		\State cut CCOMP link in dependency tree
		\State subject $\gets$ find ccomp verb's subject 
		\If{subject $\ne$ NULL and subject $\ne$ DET(e.g. `that')}
		\If{original verb do not have object}
		\State make subject as original verb's object
		\EndIf
		\ElsIf{subject $=$ DET (e.g. `that')}
		\State find root verb's object, substitute `that'
		\EndIf
		\State str1 $\gets$ traverse\_a\_string(ccomp verb)
		\State str2 $\gets$ traverse\_a\_string(original verb)
		\State \textbf{return} [str1, str2]
		\EndProcedure
	\end{algorithmic}
\end{algorithm}



\begin{algorithm}[!htb]
    \scriptsize
	\caption{Transform() in XCOMP Analyzer}\label{alg:xcomp-transform}
	\begin{algorithmic}[1]
		\Procedure{Open Clausal Component transform Procedure}{$sents$}\Comment{Input sentence}
		\State subject-token $\gets$ find\_subject(xcomp-verb-token) \Comment{find subject of the actual verb}
		\State results $\gets$ empty list
		\If{AUX $\not\in$ \emph{sents}}\Comment{for some cases two verbs needs to be output}
		\State cut xcomp link
		\State results.add(traverse\_a\_string(xcomp-verb-token))
		\EndIf
		\State remove AUX token in the dependency tree
		\State replace xcomp-verb-token in subject's children with actual-verb-token
		\State results.add(traverse\_a\_string(actual-verb-token))
		\State \textbf{return} results
		\EndProcedure
	\end{algorithmic}
\end{algorithm}


\begin{algorithm}[!htb]
    \scriptsize
	\caption{Transform() in ACL Analyzer}\label{alg:acl-transform}
	\begin{algorithmic}[1]
		\Procedure{Adjective Clause transform Procedure}{$sents$}\Comment{
		Input Sentence}
		\State cut acl edge in dependency tree
		\State str1 $\gets$ traverse\_a\_string(root-token)
		\State mid-fix $\gets$ empty string
		\If{acl-token.tag $=$ VBN and by-token in acl-token's right children}
		\State mid-fix $\gets$ `be'
		\EndIf
		\State update acl-verb-token's left children with [acl-noun, mid-fix, [t $\in$ acl-verb-token.lefts where t.dep $\not\in$ [AUX]]
		\State correct acl-verb-tense
		\State str2 $\gets$ traverse\_a\_string(acl-verb-token)
		\State \textbf{return} [str1, str2]
		\EndProcedure
		\end{algorithmic}
\end{algorithm}

\begin{algorithm}[!htb]
    \scriptsize
	\caption{Get-Temporal Function at Line 5 in Algorithm \ref{alg:adverbial-transform}}\label{alg:adverbial-temp}
	\begin{algorithmic}[1]
		\Procedure{Adverbial Clause Temporal Info Extraction Procedure}{$prep-or-mark, cur-temp$}\Comment{The PREP or MARK token in input sentence}
		\State flag $\gets$ False\Comment{whether we change the temp}
		\If{prep-or-mark.type $=$ PREP}
		\State sign $\gets$ -1
		\Else
		\State sign $\gets$ 1
		\EndIf
		\State text $\gets$ prep-or-mark.text.lower()
		\If{text $=$ \emph{as}}
		\State flag $\gets$ True
		\ElsIf{text $\in$ [\emph{until}, \emph{till}] }
		\State flag $\gets$ True
		\State cur-temp $\gets$ cur-temp + sign
		\ElsIf{text $=$ \emph{after}}
		\State flag $\gets$ True
		\State cur-temp $\gets$ cur-temp - sign
		\ElsIf{text $=$ \emph{before}}
		\State flag $\gets$ True
		\State cur-temp $\gets$ cur-temp + sign
		\EndIf
		\State \textbf{return} [flag, cur-temp]
		\EndProcedure
	\end{algorithmic}
\end{algorithm}


%% file: Appendix/appendixc.tex
\section*{Appendix C: Action Representation Fields}\label{appendix:B}
\par
Action Representation Fields (ARFs) in the demo sentence \emph{James gently throws a red ball to Alice in the restaurant from back}, extracted with SRL: 
\begin{itemize}[noitemsep]
    \small
	\item \textbf{owner}: James
	\item \textbf{target}: a red ball
	\item \textbf{prop}: to Alice
	\item \textbf{action}: throw
	\item \textbf{origin\_action}: throws
	\item \textbf{manner}: gently
	\item \textbf{modifier\_location}: in the restaurant
	\item \textbf{modifier\_direction}: from back
\end{itemize}
In this case, our output for the \emph{prop} and \emph{target} is not correct; they should be swapped. This is one example where this module can introduce errors. 
\\\\
Additional ARFs, extracted heuristically:
\begin{itemize}
    \small
	\item \textbf{startTime}: Calculated by current scene time
	\item \textbf{duration}: We have a pre-defined list of words that when appearing in the sentence, they will indicate a short duration (e.g ``run'' or ``fast'') and therefore the \emph{duration} will be set to 1 second; in contrast, for words like ``slowly'' we assign a duration of 4 seconds; otherwise, the duration is 2 seconds.
	\item \textbf{speed}: Similarly to \emph{duration}, we have pre-defined lists of words that would affect the speed of the pre-baked animation: ``angrily'' would result in faster movement, but ``carefully'' in slower movement. We have 3 scales: 0.5, 1, 2 which corresponds to \emph{slow, normal} and \emph{fast}.
	\item \textbf{translation}: We have a list of \emph{actions} which would entail a movement in from one place to another, e.g. ``go''. If the value of the \emph{action} exists in this list, it is set to True, otherwise False. 
	\item \textbf{rotation}: If the \emph{action} exists in our list of verbs that entail rotation, this field is True, otherwise False. Rotation refers to movement in place e.g. ``turn'' or ``sit''.
	\item \textbf{emotion}: We find the closet neighbor of each word in the sentence in list of words that indicate emotion, using word vector similarity. If the similarity exceeds an empirically tested threshold, then we take the corresponding emotion word as the \emph{emotion} field of this action.
	\item \textbf{partial\_start\_time}: an important field, since it controls the sequence order of each action. It determines which actions happen in parallel and which happen sequentially. This is still an open question. We solve this problem when doing sentence simplification. Together with the input sentence, current time is also fed into each \emph{Analyzer}. There are several rules in some of the \emph{Analyzers} to obtain temporal information. For example, in Line 5 of the \emph{Adverbial Clause Analyzer} (c.f.\ref{alg:adverbial-transform}), we assign different temporal sequences for simplified actions. The algorithm is shown in Algorithm~\ref{alg:adverbial-temp}. The \emph{sign} together with specific prepositions determines the change direction of current temporal id. In the \emph{Coordination Analyzer}, the current temporal id changes when it encounters two verbs sharing same subject. Then the later action will get a bigger temporal id.
\end{itemize}